\documentclass[conference, onecolumn]{IEEEtran}

\ifCLASSINFOpdf
\else
\fi

\usepackage{graphicx}

 \usepackage{subcaption}

\begin{document}
%
\title{A Deep Learning-based Detector for Brown Spot Disease in Passion Fruit Plant Leaves}



%
\author{\IEEEauthorblockN{Andrew Katumba\IEEEauthorrefmark{1},
Moses Bomera\IEEEauthorrefmark{1}, 
Cosmas Mwikirize\IEEEauthorrefmark{1},
Gorret Namulondo\IEEEauthorrefmark{1}, 
Mary Gorret Ajero\IEEEauthorrefmark{2}, \\
Idd Ramathani\IEEEauthorrefmark{2}, 
Olivia  Nakayima\IEEEauthorrefmark{1},
Grace Nakabonge\IEEEauthorrefmark{1},
Dorothy Okello\IEEEauthorrefmark{1} and
Jonathan Serugunda\IEEEauthorrefmark{1}}
\IEEEauthorblockA{\IEEEauthorrefmark{1}Department of Electrical and Computer Engineering\\
Makerere University, Kampala, Uganda\\ Corresponding Author Email: andrewkatumba@cedat.mak.ac.ug}
\IEEEauthorblockA{\IEEEauthorrefmark{2}National Crops Resources Research Institute
(NaCRRI), Kampala, Uganda\\
}
}


\maketitle


%
\IEEEpeerreviewmaketitle

\section{Introduction}
Pests and diseases pose a key challenge to passion fruit farmers across Uganda and East Africa in general. They lead to loss of investment as yields reduce and losses increases. As the majority of the farmers including passion fruit farmers, in the country are smallholder farmers from low-income households, they do not have sufficient information and means to combat these challenges. While, passion fruits have the potential to improve the well-being of these farmers given their short maturity period and high market value \cite{BECA2020}, without the required knowledge about the health of their crops, farmers can not intervene promptly to turn the situation around. 


For this work, we partnered with the Uganda National Crop Research Institute (NaCRRI) to develop a dataset of expertly labeled passion fruit plant leaves and fruits, both diseased and healthy. We made use of their extension service to collect images from five districts in Uganda to create the dataset.  Using the dataset, we are applying state-of-the-art techniques in machine learning, specifically deep learning at scale for object detection and classification for accurate plant health status prediction.

While deep learning techniques have been applied to various disease diagnosis contexts with varying degrees of success(\cite{Singh2017, Ramcharan2017, Fuentes2017, Price1993, Sambasivam2020}), there has not been any significant effort,  to the best of our knowledge, to create a dataset or apply machine learning techniques to passion fruits despite their obvious financial benefits. 

With this work, we hope to fill this gap by generating and making publically available an image dataset focusing on passion fruit plant diseases and pest damage and training the first generation of machine learning-based models for passion fruit plant disease identification using this dataset. The initial focus is on the locally prevalent woodiness (viral) and brown spot (fungal) diseases.

\section{Dataset}
As mentioned in the previous section, one of the objectives of this work is to compile a high-quality dataset of passion fruit plant fruits and leaves, starting with brown spot and woodiness diseases, on which various machine learning and computer vision tasks can be evaluated.  
The dataset was obtained in field visits to the districts of  Masaka, Buikwe, Jinja, Kayunga, and Mbale, in order to capture sufficient variation in plant features across the different regions of the country. We expect, in the short term,  to grow the dataset to over 30,000 images and make it available to the research community as an open dataset to drive further research on passion fruit diseases.
For the initial modeling using this dataset,  a subset of 2,000 images of brown spot afflicted leaves were used with a focus on the local purple passion fruit variety, as it is the predominant variety in the country.
.
\section{Methodology}
From the general dataset, 2000 images symptomatic of the brown spot disease were randomly selected for use in the training phase. The images were cropped to 400x400 pixels (to allow for larger batch sizes and quick iteration during training) and then uploaded for labeling through a convenient interface provided by the Labelbox \cite{Labelbox} platform. The labels were then exported from the platform in JSON format for use with the model training.
 
As this work was structured as an object detection task i.e. identifying brown spot patches on leaves, we made use of Convolutional Neural Networks (CNNs) as they are suitable for this computer vision task. The labels were therefore instances of bounding boxes indicating regions of the leaf exhibiting symptoms of brown spot disease.

\section{Experimental Setup and Results}
In this section, we present results for object detection with the EfficientDet \cite{EfficientDet} model. In particular, we  focused on the D3 variant of the model as it has shown competitive results on other object detection tasks. It has been demonstrated that EfficientDet models can reach high performance as shown on the COCO dataset \cite{EfficientDet} and in some cases can be an order of magnitude faster to train than other object detection models.

In parallel, we are evaluating other model classes, including YOLO\cite{Redmon2016} and SSD\cite{Liu2016} (not discussed here), to feed into a comparative study. Given the small size of the dataset, we made use of transfer learning \cite{Pan2016}, with the model pre-trained on the COCO dataset \cite{COCO2014}. The Tensorflow\cite{Tensorflow2015} machine learning framework was used for both training and inference.

The experiments were run on the Microsoft Azure Cloud on an NC6 virtual machine with 6 cores, 56 GB of RAM and 1 half an NVIDIA K80 GPU card. 90\% of the images were used to train and validate the model, while the remaining 10\% were used to test the model. 

\begin{figure}
\centering
\begin{subfigure}{.5\textwidth}
  \centering
  \includegraphics[width=0.85\linewidth]{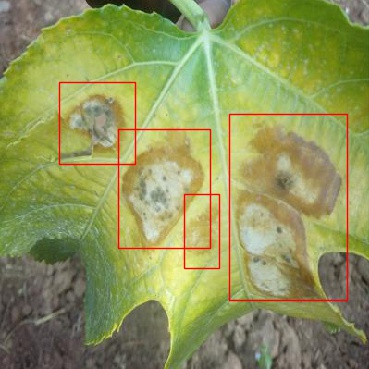}
  \caption{Groud truth}
  \label{fig:sub1}
\end{subfigure}%
\begin{subfigure}{.5\textwidth}
  \centering
  \includegraphics[width=0.85\linewidth]{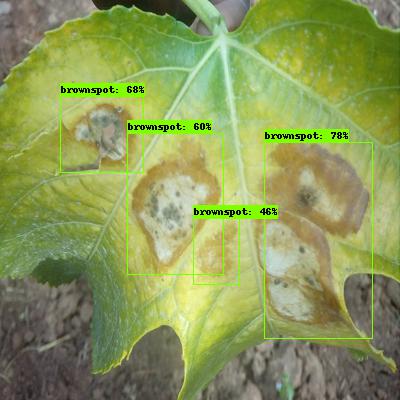}
  \caption{Model output}
  \label{fig:sub2}
\end{subfigure}
\caption{Illustration of EfficientDet D3 model test output: a)  ground truth labels and b) model  prediction. The model is able to capture all instances of brown spot disease in the image.}
\label{fig:test}
\end{figure}

We achieve a mean average precision of 0.351 on the test set after training for 10,000 iterations. An illustration of the model output for an out-of-sample example is shown in Fig. \ref{fig:test}, for which the model is able to capture all patches of brown spot disease on the leaf. Inference times are 0.06s for GPU and 0.63s for CPU.

\section{Conclusion}
Deep Learning techniques, in particular, computer vision based object detectors can be applied to capture expert knowledge for diagnosis of plant diseases. In this work, we have presented an object detection model trained on a novel dataset of passion fruit plant leaves targeting the identification and location of regions on leaves affected by the brown spot disease. Future work includes training on a larger dataset to improve the model performance and integration of the most robust model onto an edge device for use in the field by farmers and outreach workers.


\section*{Acknowledgment}

This work is being supported by a Makerere University Research Innovation Fund Grant (2019 - 2020), an NVIDIA Accelerated Data Science GPU Seed Grant and a Microsoft AI for Earth Grant (2020 - 2021). The labeling process is supported by a Labelbox Education Grant.



%

\end{document}